\pdfoutput=1

\documentclass[11pt]{article}

\usepackage[]{ACL2023}

\usepackage{times}
\usepackage{latexsym}
\usepackage{multirow}
\usepackage{adjustbox}
\usepackage{amssymb}
\usepackage{enumitem}
\usepackage{graphicx}
\usepackage{amsmath}
\usepackage{amsthm}
\usepackage{booktabs}

\usepackage[T1]{fontenc}

\usepackage[utf8]{inputenc}

\usepackage{microtype}

\usepackage{inconsolata}

\newcommand\blfootnote[1]{%
\begingroup 
\renewcommand\thefootnote{}\footnote{#1}%
\addtocounter{footnote}{-1}%
\endgroup 
}

\title{Product Question Answering in E-Commerce: A Survey}

\author{
    Yang Deng$^{1}$, Wenxuan Zhang$^{2,\dagger}$, Qian Yu$^{3}$, Wai Lam$^{4}$\\
    $^{1}$ National University of Singapore,  $^{2}$ DAMO Academy, Alibaba Group\\  $^{3}$ JD.com, $^{4}$ The Chinese University of Hong Kong\\
    \texttt{
    \{dengyang17dydy,isakzhang\}@gmail.com, yuqian81@jd.com, wlam@se.cuhk.edu.hk}
}

\begin{document}
\maketitle
\begin{abstract}
Product question answering (PQA), aiming to automatically provide instant responses to customer's questions in E-Commerce platforms, has drawn increasing attention in recent years. Compared with typical QA problems, PQA exhibits unique challenges such as the subjectivity and reliability of user-generated contents in E-commerce platforms. Therefore, various problem settings and novel methods have been proposed to capture these special characteristics. In this paper, we aim to systematically review existing research efforts on PQA. Specifically, we categorize PQA studies into four problem settings in terms of the form of provided answers. We analyze the pros and cons, as well as present existing datasets and evaluation protocols for each setting. We further summarize the most significant challenges that characterize PQA from general QA applications and discuss their corresponding solutions. Finally, we conclude this paper by providing the prospect on several future directions. 
\blfootnote{$^*$ The work described in this paper is substantially supported by a grant from the Research Grant Council of the Hong Kong Special Administrative Region, China (Project Code: 14200719).}
\blfootnote{$^\dagger$ Corresponding author.}
\end{abstract}

\section{Introduction}
\begin{figure}
\setlength{\abovecaptionskip}{5pt}   
\setlength{\belowcaptionskip}{5pt}
\centering
\includegraphics[width=0.45\textwidth]{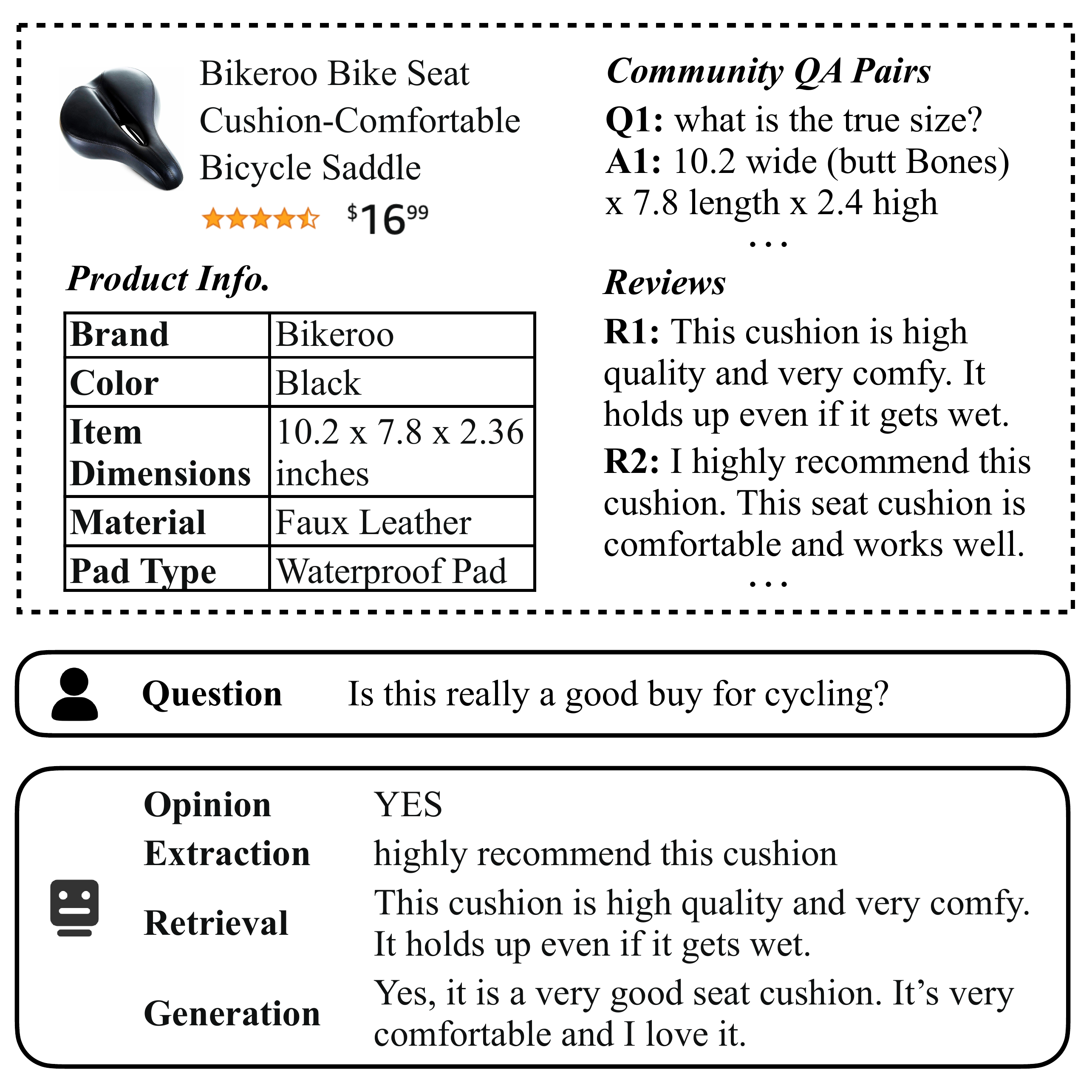}
\caption{An PQA example from Amazon.} 
\label{fig:example}
\vspace{-0.5cm}
\end{figure}

E-Commerce is playing an increasingly important role in our daily life. 
During the online shopping, potential customers inevitably have some questions about their interested products. 
To settle down their concerns and improve the shopping experience, many AI conversational assistants have been developed to solve customers' problems, such as Alexa~\cite{sigir18-pqa-challenge} and AliMe~\cite{alime}. 
The core machine learning problem underlying them, namely \textbf{Product Question Answering (PQA)}, thus receives extensive attention in both academia and industries recently. 
Figure~\ref{fig:example} depicts an actual PQA example from Amazon. 
There are a tremendous amount of product-related data available within the product page, which contains natural language user-generated content (UGC) (\textit{e.g.}, product reviews, community QA pairs), structured product-related information (\textit{e.g.}, attribute-value pairs), images, etc. 
Generally, PQA aims to automatically answer the customer-posted question in the natural language form about a specific product, based on the product-related data. 

\begin{table*}
    \centering
    \begin{adjustbox}{max width=\textwidth}
    \begin{tabular}{cccccccc}
    \toprule
    & Method & Document & Extra Data & Backbone & Main Challenge & Dataset & Pros\&Cons\\
    \midrule
    \multirow{6}{*}{Opinion}& \citet{www16-amazon-qa}&PR&-&Feature&Subjectivity&Amazon&\multirow{4}{5.1cm}{\textbf{Pro}: tackle a large proportion of questions that ask for certain opinion by using comparatively simple methods.}\\
    &\citet{icdm16-amazon}&PR&-&Feature&Subjectivity&Amazon\\
    &\citet{wsdm18-answer-pred}&PR&-&Feature&Subjectivity&Amazon\\
    &\citet{sdm19-pred}&PR&-&NN&-&Amazon\\
    &\citet{icdm19-pred}&PR&-&PLM&-&Amazon&\multirow{2}{5.1cm}{\textbf{Con}: only classify the opinion polarity without detailed info.}\\
    &\citet{naacl21-pred}&PR&QA&PLM&Low-resource&Amazon+\\
    \midrule
    \multirow{3}{*}{Extraction}
    &\citet{ijcai19-amazonqa}&PR&-&NN&Answerability&AmazonQA&\multirow{1}{5.1cm}{\textbf{Pro}: provide pinpointed answers.}\\
    & \citet{naacl19-review-mrc}&PR&MRC&PLM&Low-resource&ReviewRC&\multirow{1}{5.1cm}{\textbf{Con}: providing an incomplete answer is less user-friendly.}\\
    &\citet{emnlp20-subjqa}&PR&-&NN/PLM&Subjectivity&SubjQA\\
    \midrule
    \multirow{10}{*}{Retrieval}& \citet{acl17-demo-superagent}&PR+QA+PI&-&NN&Multi-type Resources&-&\multirow{4}{5.1cm}{\textbf{Pro}: select complete and informative sentences as the answer, based on actual customer experience.}\\
    &\citet{coling18-answer-review}&PR+QA&-&Feature&Low-resource&Amazon+\\
    &\citet{wsdm18-answer-sel}&QA&NLI&NN&Low-resource&-\\
    &\citet{www19-pqa}&PR+QA+PI&-&NN&Multi-type Resources&-\\
    &\citet{aaai19-answer-sel}&PR&QA&NN&Low-resource&Amazon+&\multirow{6}{5.1cm}{\textbf{Con}: may not answer the given question precisely since the supporting document (\textit{e.g.}, reviews) is not specifically written for answering the given question.}\\
    &\citet{kdd19-answer-sel}&PR&QA&NN&Interpretability&Amazon\\
    &\citet{sigir20-answer-sel}&QA&PR&NN&Answerability&Amazon\\
    &\citet{aacl2020}&PR+PI&QA&NN&Multi-type Resources&Amazon+\\
    &\citet{naacl21-distant-sel}&QA&CQA&PLM&Low-resource&-\\
    &\citet{www22-da-pqa}&PR&QA&PLM&Low-resource&-\\
    \midrule
    \multirow{8}{*}{Generation} 
    &\citet{wsdm19-answer-gen-chen}&PR&-&NN&-&Taobao&\multirow{4}{5.1cm}{\textbf{Pro}: provide natural forms of answers, which are specific to the given questions and flexible with different information.}\\
    &\citet{wsdm19-answer-gen-gao}&PR+PI&-&NN&Multi-type Resources&JD\\
    &\citet{oaag}&PR&-&NN&Subjectivity&Amazon\\
    &\citet{coling20-answer-gen}&PR&-&PLM&Subjectivity&AmazonQA\\
    &\citet{tois21}&PR+PI&-&NN&Multi-type Resources&JD&\multirow{4}{5.1cm}{\textbf{Con}: suffer from hallucination and factual-inconsistency issues, and lack of effective automatic evaluation methods.}\\
    &\citet{sigir21-pqa}&PR+PI&-&NN&Multi-type Resources&JD\\
    &\citet{tois22-ppqa}&PR+PI&-&NN&Personalization&Amazon\\
    &\citet{ecnlp22-semipqa}&PI&-&PLM&Multi-type Resources&semiPQA\\
    \bottomrule
    \end{tabular}
\end{adjustbox}
    \caption{Summary of PQA studies. ``Amazon+'' denotes that additional annotations are added into the ``Amazon'' dataset. ``PR'', ``QA'', and ``PI'' denote product reviews, community QA pairs, and product information, respectively.}
    \label{tab:summary}
\end{table*}

Typical QA studies~\cite{squad} and some other domain-specific QA studies (\textit{e.g.}, biomedical QA~\cite{biomedicalqa} and legal QA~\cite{legalqa}) mainly focus on the questions that ask for a certain factual and objective answer. Differently, product-related questions in PQA typically involve consumers’ opinion about the products or aspects of products. 
Therefore, early studies~\cite{icdmw11,emnlp12-answer-pred} regard PQA as a special opinion mining problem, where the answers are generated by aggregating opinions in the retrieved documents. 
Most of recent works essentially follow the same intuition, but formulate PQA as different problems in terms of the form of target answers. 
Accordingly, existing PQA studies can be categorized into four types: opinion-based, extraction-based, retrieval-based, and generation-based.
As shown in Figure~\ref{fig:example}, opinion-based PQA approaches only provide the common opinion polarity as the answer, while extraction-based PQA approaches extract specific text spans from the supporting documents as the answer. 
Retrieval-based PQA approaches further re-rank the documents to select the most appropriate one to answer the given question, while generation-based PQA approaches generate natural language sentences based on the available documents as the response. 
In this paper, we systematically review methods of these four mainstream PQA problem settings, as well as the commonly-used datasets and evaluation protocols.

Besides the task-specific challenges in each type of PQA systems, there are several common challenges across all types of PQA systems, which differentiate PQA from other QA systems. 
(1) \textbf{Subjectivity}. Subjective questions constitute a large proportion of questions in PQA, which requires to aggregate the crowd’s opinions about the questions, reflected through related reviews and QAs. 
(2) \textbf{Reliability \& Answerability}. Different from those supporting documents constructed by professionals in biomedical or legal QA, product reviews and community QA pairs come directly from non-expert users, which may suffer from some typical flaws as other UGC, such as redundancy, inconsistency, spam, and even malice. 
(3) \textbf{Multi-type resources}. The supporting documents usually consist of heterogeneous information from multi-type data resources, such as text, table, knowledge graph, image, etc. 
(4) \textbf{Low-resource}. PQA systems often encounter the low-resource issue, since different product categories may need different training data, and it is generally time-consuming and costly to manually annotate sufficient labeled data for each domain.
Accordingly, we introduce existing solutions to each challenge. 

To our knowledge, this survey is the first to focus on Product Question Answering. 
We first systematically summarize recent studies on PQA into four problem settings as well as introduce the available datasets and corresponding evaluation protocols in Section~\ref{sec:problems}. 
Then we analyze the most significant challenges that characterize PQA from other QA applications and discuss their corresponding solutions in Section~\ref{sec:challenge}. 
Finally, we discuss several promising research directions for future PQA studies and conclude this paper in Section~\ref{sec:prospect} and~\ref{sec:conclusion}.

\begin{figure*}
\setlength{\abovecaptionskip}{5pt}   
\setlength{\belowcaptionskip}{5pt}
\centering
\includegraphics[width=\textwidth]{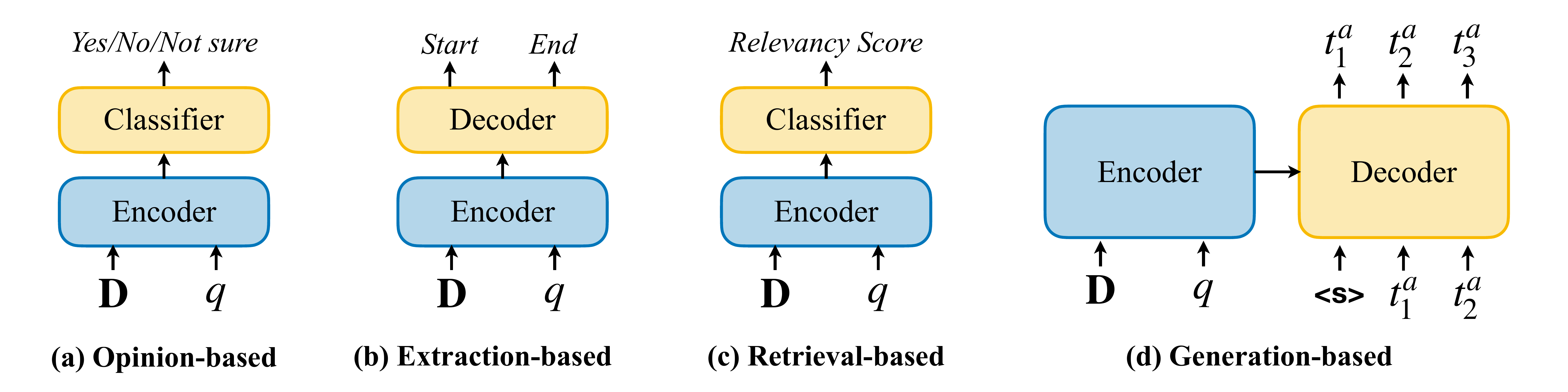}
\caption{Four main-stream problem settings in PQA studies.} 
\label{fig:paradigm}
\vspace{-0.3cm}
\end{figure*}

\section{Problems and Approaches}\label{sec:problems}
Product question answering (PQA) aims to produce an answer $a$ to a given natural language question $q$ based on a set of supporting documents $D$, where the supporting documents can be product reviews, community QA pairs, product information, etc. 
In terms of the form of provided answers, we systematically categorize the existing PQA studies into four problem settings, including Opinion-based PQA, Extraction-based PQA, Retrieval-based PQA, Generation-based PQA, and introduce corresponding approaches proposed to solve the problem, as summarized in Table~\ref{tab:summary}. 
We present an overview of the general framework for each problem setting in Figure~\ref{fig:paradigm}. In addition, the key information of the datasets adopted in existing PQA studies is summarized in Table~\ref{tab:dataset}. 

\begin{table*}
\setlength{\abovecaptionskip}{5pt}   
\setlength{\belowcaptionskip}{5pt}
    \centering
    \begin{adjustbox}{max width=\textwidth}
    \begin{tabular}{cccccccc}
    \toprule
      Dataset   & Language & Answer Form & \# Questions & \# Categories & Types of Doc. & Additional Info. & Release\\
      \midrule
      Amazon \cite{www16-amazon-qa} & English & Yes-No/Open-ended & $\sim${1.4M} & 21 & PR/PI/QA & Timestamps/User/Vote& $\checkmark$\footnotemark\\
      AmazonQA \cite{ijcai19-amazonqa} & English & Yes-No/Open-ended & $\sim${923K} & 17 & PR & Answerability & $\checkmark$\footnotemark\\
      ReviewRC \cite{naacl19-review-mrc} & English & Span & 2,596 & 2 & PR & Sentiment & $\checkmark$\footnotemark\\
      SubjQA \cite{emnlp20-subjqa} & English & Span/Open-ended & 10,098 & 6 & PR & Subjectivity & $\checkmark$\footnotemark\\
      JD \cite{wsdm19-answer-gen-gao} & Chinese & Open-ended & 469,955 & 38 & PR/PI & - & $\checkmark$\footnotemark\\
      Taobao \cite{wsdm19-answer-gen-chen} & Chinese & Open-ended & 1,155,530& 2 & PR & - & $\times$\\ 
      semiPQA \cite{ecnlp22-semipqa} & English & Open-ended & 11,243 & - & PI & - & $\times$\\
      PAGHS$^*$ \cite{ecnlp22-heter} & English & Open-ended & 309,347 & - & PR/PI/QA & Relevance of Docs. & $\times$\\
      \bottomrule
      \multicolumn{8}{l}{$^*$ PAGHS stands for Product Answer Generation from Heterogeneous Source as there is no specific name for the dataset proposed in \citet{ecnlp22-heter}. }
    \end{tabular}
    \end{adjustbox}
    \caption{Summary of existing datasets for product question answering.}
    \label{tab:dataset}
\vspace{-0.3cm}
\end{table*}

\subsection{Opinion-based PQA}
Opinion-based PQA studies focus on yes-no type questions, \textit{i.e.}, questions that can be answered by “Yes” or “No”, which constitute a large proportion on PQA platforms. 

\subsubsection{Problem Definition}
Given a product-related question $q$ and a set of supporting documents $\mathbf{D}$ (product reviews in most cases), the goal is to predict a binary answer $a\in\{\text{Yes}, \text{No}\}$. Some studies also consider the neutral answer, \textit{e.g.}, ``Not Sure".

\subsubsection{Datasets \& Evaluation Protocols}
One of the largest and widely-adopted public PQA datasets is the Amazon Product Dataset (denoted as ``Amazon'' in Table~\ref{tab:summary} and hereafter), composed by Amazon Question/Answer Data~\cite{www16-amazon-qa,icdm16-amazon} and Amazon Review Data~\cite{www16-amazon-review,emnlp19-amazon-review}. It consists of around 1.4 million answered questions and 233.1 million product reviews across over 20 different product categories. The Amazon dataset contains the information of question types (``yes-no" or ``open-ended"), answer types (``yes", ``no", or ``not sure"), helpful votes by customers, and product metadata, which is suitable for opinion-based PQA evaluation.

Due to the existence of a certain proportion of unanswerable questions based on the available reviews, it is difficult to achieve an acceptable performance with the ordinary classification accuracy metric $\text{Acc}(\mathcal{Q})$ for any method. 
Therefore, \citet{www16-amazon-qa} propose $\text{Acc@}k$, which has become the de facto metric for evaluating opinion-based PQA methods, which only calculates the classification accuracy of top-$k$ questions ranked by the prediction \textit{confidence}. 
The \textit{confidence} with each classification is its distance from the decision boundary, \textit{i.e.}, $|\frac{1}{2}-P(a|q,\mathbf{D})|$. 
A good model is supposed to assign high confidence to those questions that can be correctly addressed. 
\begin{equation} 
    \small
    \text{Acc@}k = \text{Acc}(\mathop{\arg\max}\limits_{\mathcal{Q}'\in \mathcal{P}_k(\mathcal{Q})}\sum_{q\in \mathcal{Q}'} |\frac{1}{2}-P(a|q,\mathbf{D})|)
\end{equation}
where $\mathcal{P}_k(\mathcal{Q})$ is the set of $k$-sized subsets of $\mathcal{Q}$, and $k$ is commonly set to be 50\% of the total number of questions.

\subsubsection{Methods}
\citet{www16-amazon-qa} propose a Mixtures of experts (MoEs)~\cite{moe} based model, namely Mixtures of Opinions for Question Answering (Moqa), to answer yes-no questions in PQA, where each review is regarded as an ``expert'' to make a binary prediction for voting in favor of a ``yes'' or ``no'' answer. The confidence of each review is further weighted by its relevance to the question as follows:   
\begin{equation}\label{eq:moqa}\small
    P(a|q,\mathbf{D}) = \sum_{d\in\mathbf{D}} \underbrace{P(d|q)}_\text{how relevant is $d$} \cdot \underbrace{P(a|d,q)}_\text{prediction from $d$} 
\end{equation}
Moqa is later enhanced by modeling the ambiguity and subjectivity of answers and reviews~\cite{icdm16-amazon}. 
\citet{wsdm18-answer-pred} further improve Moqa by computing the aspect-specific embeddings of reviews and questions via a three-order auto-encoder network in an unsupervised manner.
In these early studies, the features either extracted by heuristic rules or acquired from unsupervised manners may limit the performance and application of opinion-based PQA approaches. 

To better model the relation between the question and each review, \citet{sdm19-pred} and \citet{icdm19-pred} explore the utility of neural networks (\textit{e.g.}, BiLSTM~\cite{birnn}) and pretrained language models (\textit{e.g.}, BERT~\cite{bert}) to learn the distributed feature representations, which largely outperform previous methods. 
Recently, \citet{naacl21-pred} propose an approach, called SimBA (\textbf{Sim}ilarity \textbf{B}ased \textbf{A}nswer Prediction), which leverages existing answers from similar resolved questions about similar products to predict the answer for the target question.

\footnotetext[3]{\url{http://deepx.ucsd.edu/public/jmcauley/qa/}}
\footnotetext[4]{\url{https://github.com/amazonqa/amazonqa}}
\footnotetext[5]{\url{https://howardhsu.github.io/}}
\footnotetext[6]{\url{https://github.com/megagonlabs/SubjQA}}
\footnotetext[7]{\url{https://github.com/gsh199449/productqa}}

\subsubsection{Pros and Cons}
Opinion-based PQA approaches can tackle a large proportion of product-related questions that ask for certain opinion by using comparatively simple and easy-to-deploy methods. 
However, opinion-based approaches could only provide the classification result of the opinion polarity, based on the common opinion reflected in the supporting documents, without detailed and question-specific information.

\subsection{Extraction-based PQA}
Similar to typical extraction-based QA~\cite{squad} (also called Machine Reading Comprehension (MRC)), extraction-based PQA studies aim at extracting a certain span of a document to be the answer for the given product-related questions. 
\subsubsection{Problem Definition}
Given a product-related question $q$ and a supporting document $d=\{t_1,...,t_n\}\in \mathbf{D}$, which consists of one or more product reviews, the goal is to find a sequence of tokens (a text span) $a=\{t_s,...,t_e\}$ in $d$ that answers $q$ correctly, where $1 \leq s \leq n$, $1 \leq e \leq n$, and $s \leq e$.

\subsubsection{Datasets \& Evaluation Protocols}
\citet{naacl19-review-mrc} build the first extraction-based PQA dataset, called ReviewRC, using reviews from SemEval-2016 Task 5~\cite{semeval16-task5}. Similarly, \citet{ijcai19-amazonqa} conduct extensive pre-processing on the Amazon dataset~\cite{www16-amazon-qa,www16-amazon-review} to build a dataset for extraction-based PQA, called AmazonQA. It annotates each question as either answerable or unanswerable based on the available reviews, and heuristically creates an answer span from the reviews that best answer the question. 
\citet{emnlp20-subjqa} propose SubjQA dataset to investigate the relation between subjectivity and PQA in the
context of product reviews, which contains 6 different domains that are built upon TripAdvisor~\cite{tripadvisor}, Yelp\footnote{\url{https://www.yelp.com/dataset}}, and Amazon~\cite{www16-amazon-qa} datasets. 

Given the same setting as typical MRC, extraction-based PQA adopts the same evaluation metrics, including Exact Match (EM) and F1 scores. EM requires the predicted answer span to exactly match with the human annotated answer, while F1 score is the averaged F1 scores of individual answers in the token-level. 

\subsubsection{Methods}
Due to the limited training data for extraction-based PQA, \citet{naacl19-review-mrc} employ two popular pre-training objectives, \textit{i.e.}, masked language modeling and next sentence prediction, to post-train the BERT encoder on both the general MRC dataset, SQuAD~\cite{squad}, and E-Commerce review datasets, including Amazon Review~\cite{www16-amazon-review} and Yelp datasets. 
In real-world applications, there will be a large number of irrelevant reviews and the question might be unanswerable. 
To this end, \citet{ijcai19-amazonqa} first extract top review snippets for each question based on IR techniques and build an answerability classifier to identify unanswerable questions based on the available reviews. Then, a span-based QA model, namely R-Net~\cite{r-net}, is adopted for the extraction-based PQA. 
Besides, \citet{emnlp20-subjqa} develop a subjectivity-aware QA model, which performs the multi-task learning of the extraction-based PQA and subjectivity classification. Experimental results show that incorporating subjectivity effectively boosts the performance.

\subsubsection{Pros and Cons}
Extraction-based PQA approaches can provide pinpointed answers to the given questions, but it may be less user-friendly to provide an incomplete sentence to users and may also lose some additional information. Since there are a large proportion of questions that ask for certain user experiences or opinions based on the statistics in~\cite{www16-amazon-qa,tois22-ppqa}, extraction-based paradigm is less practical and favorable in real-world PQA applications. Therefore, it can be observed that there are relatively few works in extraction-based PQA studies in recent years. 

\subsection{Retrieval-based PQA}
Retrieval-based PQA studies treat PQA as an answer (sentence) selection task, which retrieves the best answer from a set of candidates to appropriately answer the given question. 
\subsubsection{Problem Definition}
Given a question $q$ and a set of supporting documents $\mathbf{D}$, the goal is to find the best answer $a$ by ranking the list of documents according to the relevancy score between the question $q$ and each document $d\in\mathbf{D}$, \textit{i.e.,} $a=\arg\max\nolimits_{d\in\mathbf{D}} \mathcal{R}(q,d)$.

\subsubsection{Datasets \& Evaluation Protocols}
Due to the absence of ground-truth question-review (QR) pairs, several efforts~\cite{aaai19-answer-sel,coling18-answer-review,aacl2020} have been made on annotating additional QR pairs into the Amazon dataset for retrieval-based PQA. 
Nevertheless, the original Amazon dataset can  be directly adopted for retrieval-based PQA studies ~\cite{www2020-answer-help,sigir20-answer-sel} that aim to select reliable or helpful answers from candidate community answers. 

Since the retrieval-based PQA methods are essentially solving a ranking problem, most studies adopt standard ranking metrics for evaluation, including mean average precision (MAP), mean reciprocal rank (MRR), and normalized discounted cumulative gain (NDCG).

\subsubsection{Methods}
\citet{acl17-demo-superagent} first demonstrate a retrieval-based PQA chatbot, namely SuperAgent, which contains different ranking modules that select the best answer from different data sources within the product page, including community QA pairs, product reviews, and  product information. 
\citet{www19-pqa} further propose a pipeline system that first classifies the question into one of the predefined question categories with a question category classifier, and then uses an ensemble matching model to rank the candidate answers. 
However, these systems usually contain multiple modules with different purposes, which require a large amount of annotated data from different sources. Therefore, most recent retrieval-based PQA works use one or two sources as the supporting documents and build the model in an end-to-end manner.

When facing a newly posted product-related question, a straight-forward answering strategy is to retrieve a similar resolved question and provide the corresponding answer to the target question. 
However, such a solution relies heavily on a large amount of domain-specific labeled data, since QA data differs significantly in language characteristics across different product categories. 
To handle the low-resource issue, \citet{wsdm18-answer-sel} propose a general transfer learning framework that adapts the shared knowledge learned from large-scale paraphrase identification and natural language inference datasets (\textit{e.g.}, Quora\footnote{\url{https://www.kaggle.com/c/quora-question-pairs}} and MultiNLI~\cite{multinli}) to enhance the performance of reranking similar questions in retrieval-based PQA systems. 
Besides, \citet{naacl21-distant-sel} propose a distillation-based distantly supervised training algorithm, which uses QA pairs retrieved by a syntactic matching system, to help learn a robust question matching model.

Another approach to obtain answers for new questions is to select sentences from product reviews. The main challenge is that the information distributions of explicit answers and review contents that can address the corresponding questions are quite different and there are no annotated ground-truth question-review (QR) pairs which can be used for training. 
\citet{coling18-answer-review} develop a distant supervision paradigm for incorporating the knowledge contained in QA collections into question-based response review ranking, where the top ranked reviews are more relevant to the QA pair and are useful for capturing the knowledge of response review ranking. 
\citet{aaai19-answer-sel} propose a multi-task deep learning method, namely QAR-net, which can exploit both user-generated QA data and manually labeled QR pairs to train an end-to-end deep model for answer identification in review data. 
\citet{kdd19-answer-sel} aim at improving the interpretability of retrieval-based PQA by identifying important keywords within the question and associating relevant words from large-scale QA pairs. 
\citet{aacl2020} employ pre-trained language models (\textit{e.g.}, BERT) to obtain weak supervision signals from the community QA pairs for measuring the relevance between the question and heterogeneous information, including natural language reviews and structured attribute-value pairs. 

For the situation where multiple user-generated answers have already been posted, \citet{sigir20-answer-sel} propose an answer ranking model, namely MUSE, which models multiple semantic relations among the question, answers, and relevant reviews, to rank the candidate answers in PQA platforms.

\subsubsection{Pros and Cons}
Retrieval-based approaches select complete and informative sentences as the answer, which may not answer the given question precisely since the supporting document (\textit{e.g.}, reviews) is not specifically written for answering the given question. 

\subsection{Generation-based PQA}
Inspired by successful applications of sequence-to-sequence (Seq2seq) models on other natural language generation tasks, several attempts have been made on leveraging Seq2seq model to automatically generate natural sentences as the answer to the given product-related question. 

\subsubsection{Problem Definition}
Given a product-related question $q$ and a set of supporting documents $\mathbf{D}$ that are relevant to the given question, the goal is to generate a natural language answer $a=\{t^a_1,t^a_2,...\}$ based on the question $q$ and supporting documents $\mathbf{D}$. 
 
 \subsubsection{Datasets \& Evaluation Protocols}
The Amazon dataset can be directly adopted for generation-based PQA. Another popular dataset used for geneartive PQA is from JD~\cite{wsdm19-answer-gen-gao}, which is one of the largest e-commerce websites in China. In total, the JD dataset contains 469,953 products and 38 product categories, where each QA pair is associated with the reviews and attributes of the corresponding product. 

Evaluating generation-based methods often involves both automatic evaluation and human evaluation. 
Common automatic evaluation metrics include (i) ROUGE~\cite{rouge} and BLEU~\cite{bleu} for evaluating lexical similarity between generated answers and ground-truth answers, (ii) Embedding-based Similarity~\cite{emb-sim}, BertScore~\cite{bertscore}, and BleuRT~\cite{bleurt} for evaluating semantic relevance, (iii) Distinct scores~\cite{dist} for evaluating the diversity of the generated answers. 
Human evaluation protocols are  designed for evaluating different perspectives of the generated answer by human annotations, such as fluency, consistency, informativeness, helpfulness, etc. 
 
\subsubsection{Methods}
Generation-based PQA studies typically regard the retrieval of relevant documents as a pre-processing step, and build the method upon the retrieved documents. 
Due to the noisy nature of retrieved documents, \citet{wsdm19-answer-gen-gao} employ a Wasserstein distance based adversarial learning method to denoise the irrelevant information in the supporting reviews, while \citet{wsdm19-answer-gen-chen} design an attention-based weighting strategy to highlight the relevant words appearing in the retrieved review snippets. 
Besides identifying relevant information from the retrieved documents, \citet{oaag} find that the rich personal opinion information in product reviews also attaches great importance in generation-based methods, as there are a large number of subjective questions in PQA. To this end, a joint learning model of answer generation and opinion mining is proposed to generate opinion-aware answers. 
Likewise, \citet{coling20-answer-gen} propose a cross-passage hierarchical memory network to  identify the most prominent opinion across different reviews for answer generation in PQA. 

Some recent works focus on leveraging documents from multi-type resources to generate the answer. \citet{sigir21-pqa} model the logical relation between unstructured documents (reviews) and structured documents (product attributes) with a heterogeneous graph neural network. \citet{tois21} aim at solving the safe answer problem  during the generation (\textit{i.e.}, neural models tend to generate meaningless and general answers), by systematically modeling product reviews, product attributes, and answer prototypes. \citet{ecnlp22-semipqa} propose present the semiPQA dataset to benchmark PQA over semi-structured data.

\subsubsection{Pros and Cons}
Generation-based methods can provide natural forms of answers specific to the given questions. However, the hallucination and factual inconsistency issues are prevalent in generation-based methods. In addition, it is still lack of robust automatic evaluation protocols for generation-based methods. 

\section{Challenges and Solutions}\label{sec:challenge}
Although the aforementioned PQA methods are developed based on different problem settings, there are some common challenges in PQA, as presented in Table~\ref{tab:summary}. Several main challenges and their corresponding solutions are summarized as follows.

\subsection{Subjectivity}
Different from typical QA whose answers are usually objective and unique, a large proportion of questions in PQA platforms are asking for subjective information or opinions. 
Meanwhile, the UGC in E-commerce such as product reviews also provides rich information about other customers' opinion. 
Therefore, early studies regard PQA as a special opinion mining problem~\cite{icdmw11,emnlp12-answer-pred}, which is followed by recent opinion-based PQA studies~\cite{www16-amazon-qa,icdm16-amazon}. 
Ideal answers to this kind of questions require information describing personal opinions and experiences. 
There are two specific challenges in exploiting such subjective information to facilitate PQA:
\begin{itemize}[leftmargin=*]
    \item \textbf{Detect question-related opinion}. 
    A common solution is to regard the question as the target aspect for aspect-based opinion extraction. 
    For example, \citet{emnlp20-subjqa} use OpineDB~\cite{opineDB} and some syntactic extraction patterns to extract opinion spans. 
    \citet{oaag} employ a dual attention mechanism to highlight the question-related information in reviews for the joint learning with an auxiliary opinion mining task. 
    \citet{findings21-absaqa} study aspect-based sentiment analysis in PQA, which classifies the sentiment polarity towards certain product aspects in the question from the community answers. 
    \item \textbf{Aggregate diverse opinion information}. 
    Since users may differ in opinions towards the same question, a good PQA system should avoid expressing a random opinion, or even being contradictory to the common opinion. 
    To this end, \citet{oaag} employ an opinion self-matching layer and design two kinds of opinion fusion strategies to uncover the common opinion among multiple reviews for generation-based PQA. Likewise, \citet{coling20-answer-gen} propose a cross-passage hierarchical memory network to identify the most prominent opinion. 
    However, existing studies pay little attention on resolving conflicting user opinions, which is a common issue in opinion summarization of product reviews~\cite{acl18-opinionsum,acl20-opinionsum} and worth exploring in the future studies of PQA. 
\end{itemize}

\subsection{Answer Reliability \& Answerability}
Similar to other UGC, product reviews and community answers in E-commerce sites, which are also provided by online users instead of professionals, vary significantly in their qualities and inevitably suffer from some reliability issues such as spam, redundancy, and even malicious content. 
Therefore, it is of great importance to study the answer reliability and answerability issue when building automatic PQA systems using these UGC. 
In terms of the availability of candidate answers, existing solutions can be categorized into two groups: 
\begin{itemize}[leftmargin=*]
    \item \textbf{Reliability of user-generated answers}. 
    When there are a set of candidate user-generated answers for the concerned question, the reliability measurement of these answers has been investigated from different perspectives. 
    For example, \citet{www2020-answer-help} predict the helpfulness of user-generated answers by investigating the opinion coherence between the answer and crowds’ opinions reflected in the reviews, while \citet{emnlp20-answerfact} tackle the veracity prediction of the user-generated answers for factual questions as an evidence-based fact checking problem. 
    However, these studies mainly focus on the content reliability while neglecting the reliability degree of the answerer~\cite{wsdm17-reliab,kdd20-reliab}.
    \item \textbf{Unanswerable questions based on the available documents}. 
    Question answerability detection has drawn extensive attention in typical QA studies~\cite{squad2}. 
    Similarly, \citet{ijcai19-amazonqa} train an binary classifier to classify the question answerability for PQA. 
    \citet{ecml20-unreliabelreview} propose a conformal prediction based framework to reject unreliable answers and return \textit{nil} answers for unanswerable questions. 
    Meanwhile, the answerablity in PQA is also highly related to the reliability of product reviews~\cite{ecnlp22-genpqa,ecnlp22-heter}. 
\end{itemize}

\subsection{Multi-type Resources}
Another characteristic of PQA is the necessity of processing heterogeneous information from multi-type resources, including natural language UGC (\textit{e.g.}, reviews, community QA pairs), structured product information (\textit{e.g.}, attribute-value pairs~\cite{econlp18,product-specification}, knowledge graph~\cite{ccks19-kbqa}), E-manuals~\cite{emanual}, images, etc. 
Early works~\cite{acl17-demo-superagent,www19-pqa} design separated modules to handle the questions that require different types of data resources. 
However, these PQA systems rely heavily on annotated data from different types of resources and neglect the relation among heterogeneous data. 
Therefore, some recent studies focus on manipulating heterogeneous information from multi-type resources in a single model for better answering product-related questions. 
For instance, \citet{aacl2020} design a unified heterogeneous encoding scheme that transforms structured attribute-value pairs into a pesudo-sentence. 
\citet{wsdm19-answer-gen-gao} employ a key-value memory network to store and encode product attributes for answer decoding with the encoded review representations, which is further combined with answer prototypes~\cite{tois21}. 
\citet{sigir21-pqa} propose a heterogeneous graph neural network to track the information propagation among different types of information for modeling the relational and logical information. 

\subsection{Low-resource}
Since there are a large amount of new questions posted in PQA platforms every day and the required information to answer the questions varies significantly across different product categories (even across different single products), traditional supervised learning methods become data hungry in this situation. However, it is time-consuming and labor-intensive to obtain sufficient domain-specific annotations. 
Existing solutions typically leverage external resources to mitigate the low-resource issue. 
In terms of the external resources, these solutions can be categorized into two groups: 
\begin{itemize}[leftmargin=*]
    \item \textbf{Transfer learning from out-domain data}. This group of solutions typically leverages large-scale open-domain labeled datasets and design appropriate TL strategy for domain adaptation in PQA. For example, \citet{wsdm18-answer-sel} transfer the knowledge learned from Quora and MultiNLI datasets to retrieval-based PQA models, by imposing a regularization term on the weights of the output layer to capture both the inter-domain and the intra-domain relationships. 
    \citet{naacl19-review-mrc} perform post-training on the SQuAD dataset to inject  task-specific knowledge into BERT for extraction-based PQA. 
    \item \textbf{Distant supervision from in-domain data}. Another line of solutions adopt the resolved QA pairs from similar products~\cite{naacl21-pred} or products in the same categories~\cite{coling18-answer-review,aaai19-answer-sel,kdd19-answer-sel,www22-da-pqa} as weak supervision signals. For example, \citet{aacl2020} and \citet{naacl21-distant-sel} employ syntactic matching systems (\textit{e.g.}, BM25) or pre-trained text embeddings (\textit{e.g.}, BERT) to obtain resolved QA pairs for facilitating the distantly supervised training process. 
\end{itemize}

\section{Prospects and Future Directions}\label{sec:prospect}
Considering the challenges summarized in this paper, we point out several promising prospects and future directions for PQA studies: 
\begin{itemize}[leftmargin=*]
    \item \textbf{Question Understanding}. 
    Due to the diversity of product-related questions, some attempts have been made on identifying the user's intents~\cite{ictir18-question}, the question types~\cite{acl17-demo-superagent}, and even the user's purchase-state~\cite{sigir21-question} from the questions. 
    In addition, some researches investigate the user's uncertainty or the question's ambiguity towards the product by asking clarifying questions~\cite{naacl21-clari,www21-clari}. 
    Despite the extensive studies for QA, question understanding has not been deeply studied in the context of PQA. 
    For example, the system should be capable of identifying the subjectivity from the product-related questions~\cite{emnlp20-subjqa}, such as opinionated questions~\cite{oaag}, comparative questions~\cite{wsdm22-compare-q}, etc. 
    \item \textbf{Personalization}. 
    As mentioned before, compared with typical QA studies~\cite{squad}, there is a large proportion of subjective questions~\cite{www16-amazon-qa} on PQA platforms, which involve user preference or require personal information to answer, rather than objective or factoid questions that look for a certain answer.
    Besides, in E-Commerce, different customers often have certain preferences over product aspects or information needs~\cite{kdd19-description,www19-tips}, leading to various expectations for the provided answers.
    Therefore, \citet{sigir18-pqa-challenge} state that a good PQA system should answer the customer’s questions with the context of her/his encounter history, taking into consideration her/his preference and interest. 
    Such personalization can make the answer more helpful for customers and better clarify their concerns about the product~\cite{tois22-ppqa}. 
    \item \textbf{Multi-modality}. 
    Compared with the widely-studied natural language UGC and structured product knowledge data, image data has received little attention in PQA studies. 
    On E-Commerce sites, there exist not only a great number of official product images, but also increasing user-shared images about their actual experiences, which  benefit many other E-Commerce applications~\cite{acl21-multimodal-help,emnlp20-multimodal}.
    The multimodal data can provide more valuable and comprehensive information for PQA systems. 
    \item \textbf{Datasets and Benchmarks}. Despite the increasing attentions on developing PQA systems, the publicly available resources for PQA are still quite limited. 
    Most existing PQA studies are evaluated on the Amazon dataset~\cite{www16-amazon-qa}, which is directly crawled from the Amazon pages. 
    Some researches \cite{ecnlp22-genpqa,ecnlp22-heter} have discussed several drawbacks of evaluating PQA systems on this dataset: 1) The ground-truth answers are quite noisy, since they are the top-voted community answers posted by non-expert users. 2) There are no annotations for assessing the relevance of the supporting documents, which may cast potential risks on the reliability of the PQA systems. 
    To facilitate better evaluations, many other data resources for PQA studies have been constructed as presented in Table~\ref{tab:dataset}. 
    However, due to the privacy or the commercial issues, some of the datasets cannot be publicly released. 
    Therefore, there is still a great demand for a large-scale, high-quality, and publicly available benchmark dataset for the future studies on PQA.  

    \item \textbf{Evaluation Protocols}. 
    The types of questions vary in a wide range, from yes-no questions to open-ended questions~\cite{www16-amazon-qa}, from objective questions to subjective questions~\cite{emnlp20-subjqa}, from factual questions to non-factual questions~\cite{emnlp20-answerfact}. 
    Different types of questions may involve different specific evaluation protocol. 
    For example, it is necessary to evaluate the precision of opinion in the answers for subjective questions~\cite{oaag}, while the veracity or factualness is important in factual questions~\cite{emnlp20-answerfact}. 
    Especially for generation-based PQA methods, the evaluation is still largely using lexical-based text similarity metrics, which are not correlated well with human judgements. 
    
\end{itemize}

\section{Conclusions}\label{sec:conclusion}
This paper makes the first attempt to overview recent advances on PQA. We systematically categorize recent PQA studies into four problem settings, including Opinion-based, Extraction-based, Retrieval-based, and Generation-based, and summarize the existing methods and evaluation protocols in each category. We also analyze the typical challenges that distinguish PQA from other QA studies. Finally, we highlight several potential directions for facilitating future studies on PQA. 

\section*{Limitations}
Since product question answering (PQA) is actually a domain-specific application in general QA, the scope of the problem may be limited. However, in recent years, PQA has received increasing attention in both academy and industry. (1) From the research perspective, PQA exhibits some unique characteristics and thus brings some interesting research challenges as discussed in Section~\ref{sec:challenge}. For example, some studies use PQA as an entrypoint to analyze the subjectivity in QA tasks.  (2) From the application perspective, it has great commercial value. Online shopping is playing an increasingly important role in everyone’s daily life, so that many high-tech companies develop AI conversational assistants for promptly solving customer’s online problems, including but not limited to Amazon, eBay, Alibaba, JD, etc. 
Regarding the large amount of research efforts that have been made, there is not a systematic and comprehensive review about this research topic. 
Similar to recent surveys of other domain-specific QA, such as biomedical QA~\cite{biomedicalqa} and legal QA~\cite{legalqa}, we hope that this paper can serve as a good reference for people working on PQA or beginning to work on PQA, as well as shed some light on future studies on PQA and raise more interests from the community for this topic.

\bibliography{custom}
\bibliographystyle{acl_natbib}

\end{document}